\documentclass[11pt,a4paper]{article}
\usepackage[hyperref]{acl2021}
\usepackage{times}
\usepackage{latexsym}

\usepackage{microtype}

\aclfinalcopy 

\usepackage{amsmath}
\usepackage{algorithm}
\usepackage{algorithmic}
\usepackage{booktabs}
\usepackage{multirow}
\usepackage{tabularx}
\usepackage{amsfonts}
\usepackage{graphicx}

\title{Continual Learning for Task-oriented Dialogue System with Iterative Network Pruning, Expanding and Masking}

\author{Binzong Geng\textsuperscript{\rm 1,2}\thanks{\ \;This work was conducted when Binzong Geng was an intern at SIAT, Chinese Academy of Sciences.},~ Fajie Yuan\textsuperscript{\rm 3},~Qiancheng Xu\textsuperscript{\rm 4},~Ying Shen\textsuperscript{\rm 5},~Ruifeng Xu\textsuperscript{\rm 6},~Min Yang\textsuperscript{\rm 2}\thanks{\ \;Min Yang is corresponding author.} \\
\textsuperscript{\rm 1}University of Science and Technology of China\\
\textsuperscript{\rm 2}Shenzhen Institute of Advanced Technology, Chinese Academy of Sciences \\
\textsuperscript{\rm 3}Westlake University \quad
\textsuperscript{\rm 4}Georgia Institute of Technology\\
\textsuperscript{\rm 5}Sun Yat-sen University \quad
\textsuperscript{\rm 6}Harbin Institute of Technology (Shenzhen)\\
\texttt{\{bz.geng,min.yang\}@siat.ac.cn}, \texttt{yuanfajie@westlake.edu.cn} \\ 
\texttt{qxu309@gatech.edu}, \texttt{sheny76@mail.sysu.edu.cn}, \texttt{xuruifeng@hit.edu.cn}
}

\date{}

\begin{document}
\maketitle
\begin{abstract}
This ability to learn consecutive tasks without forgetting how to perform previously trained problems is essential for developing an online dialogue system. This paper proposes an effective continual learning for the \underline{t}ask-oriented dialogue system with iterative network \underline{p}runing, \underline{e}xpanding and \underline{m}asking (TPEM), which preserves performance on previously encountered tasks while accelerating learning progress on subsequent tasks. Specifically, TPEM (i) leverages network pruning to keep the knowledge for old tasks, (ii) adopts network expanding to create free weights for new tasks, and (iii) introduces task-specific network masking to alleviate the negative impact of fixed weights of old tasks on new tasks. We conduct extensive experiments on seven different tasks from three benchmark datasets and show empirically that TPEM leads to significantly improved results over the strong competitors.  For reproducibility, we submit the code and data at: \textcolor{blue}{\url{https://github.com/siat-nlp/TPEM}}.

\end{abstract}

\section{Introduction}
Building a human-like task-oriented dialogue system is a long-term goal of AI. Great endeavors  have been made
in designing end-to-end task-oriented dialogue systems (TDSs) with sequence-to-sequence (Seq2Seq) models \cite{eric2017copy,madotto2018mem2seq,gangi-reddy-etal-2019-multi,qin2020dynamic,mi2019meta,he2020amalgamating,wang2020dual,qin2021exploring}, which have taken the state-of-the-art of TDSs to a new level. Generally, Seq2Seq models leverage an encoder to create a vector representation of dialogue history and KB information, and then 
pass this representation into a decoder so as to output a response word by word. For example, GLMP \cite{DBLP:journals/corr/abs-1901-04713} is a representative end-to-end TDS, which incorporates KB  information into Seq2Seq model by using a global memory  pointer to filter irrelevant KB knowledge  and a local memory pointer to instantiate entity slots.


Despite the remarkable progress of previous works, the current dominant paradigm for TDS
is to learn a Seq2Seq model on a given dataset specifically for a particular purpose, which is referred to as isolated learning.
Such learning paradigm is theoretically of limited success in accumulating the knowledge it has learned before. When a stream of domains or functionalities are joined to be trained sequentially, isolated learning
faces catastrophic forgetting \cite{mccloskey1989catastrophic,yuan2020parameter,yuan2020one}.  In contrast, humans retain and accumulate knowledge throughout their lives so that they become more efficient and versatile facing new tasks in future learning \cite{thrun1998lifelong}. If one desires to create a human-like dialogue system, imitating such a lifelong learning skill is quite necessary. 

This paper is motivated by the fact that a cognitive AI has continual learning ability by nature to develop a task-oriented dialogue agent that can accumulate knowledge learned in the past and use it seamlessly in new domains or functionalities. 
Continual learning \cite{parisi2019continual,wu2018memory,yuan2020parameter,yuan2020one} is hardly a new idea for machine learning, but remains as a non-trivial step for building empirically successful AI systems. It is essentially the case for creating a high-quality TDS. On the one hand, a dialogue system is expected to reuse previously acquired knowledge, but focusing too much on stability may hinder a TDS from quickly adapting to a new task. On the other hand, when a TDS pays too much attention to plasticity, it may quickly forget previously-acquired abilities \cite{mallya2018packnet}.

In this paper, we propose a continual learning method for \underline{t}ask-oriented dialogue system with iterative network \underline{p}runing, \underline{e}xpanding and \underline{m}asking (TPEM), which preserves performance on previously encountered tasks while accelerating learning progress on the future tasks.
Concretely, TPEM adopts the global-to-local memory pointer networks (GLMP) ~\cite{DBLP:journals/corr/abs-1901-04713} as the base model due to its powerful performance in literature and easiness for implementation. We leverage iterative pruning to keep old tasks weights and thereby avoid forgetting. Meanwhile, a network expanding strategy is devised to gradually create free weights for new tasks. Finally, we introduce a task-specific binary matrix to mask some old task weights that may hinder the learning of new tasks. It is noteworthy that TPEM is model-agnostic since the pruning, expanding and binary masking mechanisms merely work on weight parameters (weight matrices) of GLMP. 

We conduct extensive experiments on seven different domains from three benchmark TDS datasets. Experimental results demonstrate  that our TPEM method significantly outperforms strong baselines for task-oriented dialogue generation in continual learning scenario.

\section{Our Methodology}

\subsection{Task Definition}
Given the dialogue history $X$ and KB tuples $B$, TDS aims to generate the next system response $Y$ word by word. Suppose a lifelong TDS model that can handle domains 1 to $k$ has been built, denoted as $\mathcal{M}_{1:k}$. The goal of TDS in continual learning scenario is to train a model $\mathcal{M}_{1:k+1}$ that can generate responses of the $k+1$-th domain without forgetting how to generate responses of previous $k$ domains. We use the terms ``domain'' and ``task'' interchangeably, because each of our tasks is from a different dialogue domain.




\subsection{Overview}
In this paper, we adopt the global-to-local memory pointer networks (GLMP)~\cite{DBLP:journals/corr/abs-1901-04713} as base model, which has shown powerful performance in TDS. We propose a continual learning method for TDS with iterative pruning, expanding, and masking. In particular, we leverage pruning to keep the knowledge for old tasks. Then, we adopt network expanding to create free weights for new tasks. Finally, a task-specific binary mask is adopted to mask part of old task weights, which may hinder the learning of new tasks. 
The proposed model is model-agnostic since the pruning, expanding and binary masking mechanisms merely work on weight parameters (weight matrices) of the encoder-decoder framework. 
Next, we will introduce each component of our TPEM framework in detail. 

 
\subsection{Preliminary: The GLMP Model}
GLMP contains three primary components: external knowledge, a global memory encoder, and a local memory decoder. Next, we will briefly introduce the three components of GLMP. 
The readers can refer to \cite{DBLP:journals/corr/abs-1901-04713} for the implementation details. 

\paragraph{External Knowledge}
To integrate external knowledge into the Seq2Seq model, GLMP adopts the end-to-end memory networks to encode the word-level information for both dialogue history (dialogue memory) and structural knowledge base (KB memory). Bag-of-word representations are utilized as the memory embeddings for two memory modules. 
Each object word is copied directly when a memory position is pointed to. 

\paragraph{Global Memory Encoder}
We convert each input token of dialogue history into a fixed-size vector via an embedding layer. The embedding vectors go through a bi-directional recurrent unit (BiGRU) \cite{chung2014empirical} to learn contextualized dialogue representations. The original memory representations and the corresponding implicit representations will be summed up, so that these contextualized representations can be written into the dialogue memory. Meanwhile, the last hidden state of dialogue representations is used to generate two outputs (i.e.,  global memory pointer and  memory readout) by reading out from the external knowledge. 
Note that an auxiliary  multi-label classification task is added to train the global memory pointer as a multi-label classification task. 

\paragraph{Local Memory Decoder}
Taking the global memory pointer, encoded dialogue history and KB knowledge as input, a sketch GRU is applied to generate a sketch response $Y^s$  that includes the sketch tags rather than slot values. If a sketch tag is generated, the global memory pointer is then passed to the external knowledge and the retrieved object word will be picked up by the local memory pointer; otherwise, the output word is generated by the sketch GRU directly.  

To effectively transfer knowledge for subsequent tasks and reduce the space consumption, the global memory encoder and external knowledge in GLMP are shared among all tasks, while a separate local memory decoder is learned by each task.

 \subsection{Continual Learning for TDS}
 We employ an iterative network pruning, expanding and masking framework for TDS in continual learning scenario, inspired by \cite{mallya2018packnet,mallya2018piggyback}.
\paragraph{Network Pruning}
To avoid  ``catastrophic forgetting''  of GLMP, a feasible way is to retain the acquired old-task weights and enlarge the network by adding  weights for learning new tasks. However, as the number of tasks grows, the complexity of model architecture increases rapidly,
 making the deep model difficult to train. To avoid constructing a huge network, we compress the model for the current task by releasing a certain fraction of neglectable weights of old tasks ~\cite{DBLP:journals/corr/abs-1803-03635,geng2021iterative}.

Suppose that for task $k$, a compact model $\mathcal{M}_{1:k}$ that is able to deal with tasks 1 to $k$ has been created and available. We then free up a certain fraction of neglectable weights (denoted as $\mathbf{W}^F_k$) that have the lowest absolute weight values by setting them to zero. The released weights associated with task $k$ are extra weights which can be utilized repeatedly for learning newly coming tasks. However, pruning a network suddenly changes the network connectivity and thereby leads to performance deterioration. To regain its original performance after pruning, we re-train the preserved weights for a small number of epochs. After a period of pruning and re-training,
we obtain a sparse network with minimal performance loss on the performance of task $k$. This network pruning and re-training procedures are performed iteratively for learning multiple subsequent tasks. When inferring task $k$, the released weights are masked in a binary on/off fashion such that the network state keeps consistent with the one learned during training. 


\paragraph{Network Expanding}
The amount of preserved weights for old tasks becomes larger with the growth of new tasks, and there will be fewer free weights for learning new tasks, resulting in slowing down the learning process and making the found solution non-optimal. 
An intuitive solution is to expand the model while learning new tasks so as to increase new capacity of the GLMP model for subsequent tasks~\cite{10.1145/3323873.3325053,hung2019compacting}. 

To effectively perform network expansion while keeping the compactness of network architecture, we should consider two key factors: (1) the proportion of free weights for new tasks (denoted as $F_k$) and (2) the number of training batches (denoted as $N_k$). Intuitively, it is difficult to optimize the parameters that are newly added and randomly initialized  with a small number of training data. To this end, we define the following strategy to expand the hidden size $H_k$ for the $k$-th task from  $H_{k-1}$:
\begin{equation}
\label{eq:expand}
\resizebox{0.87\hsize}{!}{%
$H_k = H_{k-1} + \alpha  * (P_{k-1} - F_k)
* \log(1+N_k/\beta)$}
\end{equation}
where $\alpha$ and $\beta$ are two hyperparameters. $P_{k-1}$ is the pruning ratio of task $k-1$. In this way, we are prone to expand more weights for the tasks that have less free weights but more training data.



\begin{table*}[ht!]
  \centering
  {\renewcommand{\arraystretch}{1.0}
\resizebox{2.0\columnwidth}{!}{
    \begin{tabular}{ccccccccc}
    \toprule
    Task ID & 1     & 2     & 3     & 4     & 5     & 6     & 7     &  \\
    \midrule
    Task  & Schedule & Navigation & Weather & Restaurant & Hotel & Attraction & CamRest & Avg. \\
    \midrule
    Ptr-Unk & 0.00/23.33 & 0.36/14.17 & 1.26/12.62 & 1.20/21.21 & 1.66/16.14 & 0.84/19.16 & 8.40/39.45 & 1.96/20.87 \\
    Mem2Seq & 0.66/23.32 & 3.87/23.37 & 3.21/38.90 & 1.37/14.17 & 0.95/10.25 & 0.19/4.80 & 10.10/43.07 & 2.91/22.55 \\
    GLMP  & 0.95/15.01 & 3.91/24.34 & 2.56/27.12 & 6.51/32.76 & 5.24/29.60 & 6.72/30.31 & 16.96/\textbf{52.85} & 6.12/30.28 \\
    UCL   & 12.60/60.24 & 4.42/33.06 & 4.27/47.93 & 3.57/15.60 & 2.40/10.34 & 1.20/14.24 & 12.77/39.74 & 5.89/31.59 \\
    Re-init & 16.21/64.06 & 9.38/42.47 & 11.54/50.30 & 8.97/\textbf{34.06} & 6.52/\textbf{33.60} & 3.78/18.05 & 16.88/48.15 & 10.47/41.53 \\
    Re-init-expand & 15.98/64.29 & 9.92/40.15 & 11.50/54.12 & \textbf{9.41}/30.98 & 6.07/31.54 & 5.80/17.56 & 16.60/46.42 & 10.75/40.72 \\
    \midrule
    TPEM  & \textbf{16.72}/\textbf{67.15} & \textbf{11.95}/\textbf{49.74} & \textbf{13.27}/\textbf{55.60} & 7.98/31.90 & \textbf{7.07}/30.99 & \textbf{9.11}/\textbf{33.74} & \textbf{17.60}/51.77 & \textbf{11.96}/\textbf{45.84} \\
    w/o Pruning & 16.68/66.74 & 11.33/45.01 & 13.07/51.76 & 7.67/30.02 & 6.57/33.25 & 8.96/23.56 & 17.48/52.08 & 11.68/43.20 \\
    w/o Expansion & \textbf{16.72}/\textbf{67.15} & \textbf{11.95}/\textbf{49.74} & 11.35/51.85 & 7.40/31.73 & 5.17/32.89 & 8.71/29.63 & 15.17/52.16 & 10.92/45.02 \\
    w/o Masking & \textbf{16.72}/\textbf{67.15} & 11.35/48.48 & 11.88/54.25 & 7.29/31.79 & 6.21/32.59 & 8.42/30.78 & 16.71/51.35 & 11.23/45.20 \\
    \bottomrule
    \end{tabular}}}
    \caption{BLEU/Entity F1 results evaluated on the final model after all 7 tasks are visited. We use Avg. to represent the average performance of all tasks for each method.}  
  \label{table2}%
\end{table*}%

\paragraph{Network Masking}
The preserved weights $\mathbf{W}_k^P$ of old tasks are fixed so as to retain the performance of learned tasks and avoid forgetting. However, not all preserved weights are beneficial to learn new tasks, especially when there is a large gap between old and new tasks. 
To resolve this issue, we apply a learnable binary mask $\mathbf{M}^k$ for each task $k$ to filter some old weights that may hinder the learning of new tasks.  We additionally maintain a matrix $\tilde{\mathbf{M}}^k$ of real-valued mask weights, which has the same size as the weight matrix $\mathbf{W}$. The binary mask matrix $\mathbf{M}^k$, which participates in forward computing, is obtained by passing each element of $\tilde{\mathbf{M}}^k$ through a binary thresholding function:
\begin{equation}
\mathbf{M}^k_{ij} =
\begin{cases} 
1,  & \mbox{if ~~}\tilde{\mathbf{M}}^k_{ij}>\tau \\
0, & \mbox{ortherwise }
\end{cases}
\end{equation}
where $\tau$ is a pre-defined threshold. The real-valued mask $\tilde{\mathbf{M}}^k$ will be updated in the backward pass via gradient descent. After obtaining the binary mask $\mathbf{M}^k$ for a given task, we discard  $\tilde{\mathbf{M}}^k$ and only store $\mathbf{M}^k$.  The weights selected are then
represented as $\mathbf{M}^k \odot \mathbf{W}_k^P$, which get along with free weights $\mathbf{W}_k^F$ to learn new tasks. Here, $\odot$ denotes element-wise product. Note that old weights $\mathbf{W}^P_k$ are ``picked'' only and keep unchanged during training. Thus, old tasks can be recalled without forgetting.
Since a binary mask  requires only one extra bit per parameter,  
TPEM only introduces an approximate overhead of 1/32 of the backbone network size per parameter, given that a typical network parameter is often represented by a 32-bit float value.

\section{Experimental Setup}
\paragraph{Datasets}
Since there is no authoritative dataset for TDS in continual learning scenario, we evaluate  TPEM  on 7 tasks from three benchmark TDS datasets: (1) In-Car Assistant~\cite{eric2017copy}
that contains 2425/302/304 dialogues for training/validation/testing, belonging to calendar scheduling, weather query, and POI navigation domains, (2) Multi-WOZ 2.1~\cite{budzianowski2018multiwoz} that contains 1,839/117/141 dialogues for training/validation/testing, belonging to restaurant, attraction, and hotel domains, and (3) CamRest~\cite{DBLP:journals/corr/WenGMRSUVY16a} that contains 406/135/135 dialogues from the  restaurant reservation domain for training/validation/testing.

\paragraph{Implementation Details~~}
Following~\cite{DBLP:journals/corr/abs-1901-04713}, the word embeddings are randomly initialized from normal distribution $\mathcal{N}(0,0.1)$ with size of 128. 
We set the size of encoder and decoder as 128.
We conduct one-shot pruning with ratio $P=0.5$. The hyperparameters $\alpha$ and $\beta$ are set to 32 and 50, respectively. 
We use Adam optimizer to train the model, with an initial learning rate of $1e^{-3}$. The batch size is set to 32 and the number of memory hop $k$ is set to 3.
We set the maximum re-training epochs to 5. That is, we adopt the same re-training epochs for different tasks.
We run our model three times and report the average results. 

\paragraph{Baseline Methods~~}
First, we compare TPEM with three widely used TDSs: \textbf{Ptr-Unk}~\cite{eric2017copy},
\textbf{Mem2Seq}~\cite{madotto2018mem2seq},
and \textbf{GLMP}~\cite{DBLP:journals/corr/abs-1901-04713}. 
In addition, we also compare TPEM with \textbf{UCL}~\cite{ahn2019uncertainty} which is a popular continual learning method. Furthermore, we report results obtained by the base model when its parameters are optionally re-initialized after a task has been visited (denoted as \textbf{Re-init}). We also report the results of Re-init with network expansion (denoted as \textbf{Re-init-expand}). Different from GLMP that keeps learning a TDS by utilizing parameters learned from past tasks as initialization for the new task, both Re-init and Re-init-expand save a separate model for each task in inference without considering the continual learning scenario.

\begin{figure*}[t]
    \centering
    \includegraphics[width=2\columnwidth]{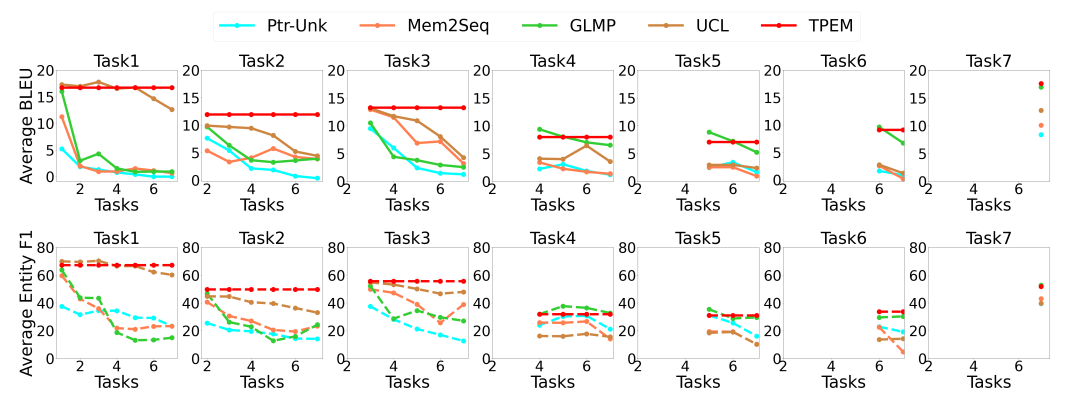}
    \caption{The change of BLEU/Entity F1 scores for each task during the whole learning process (i.e., after learning new tasks).}
    \label{fig:task1_7}
\end{figure*}

\begin{figure}[t]
  \centering
  \includegraphics[width=1\columnwidth]{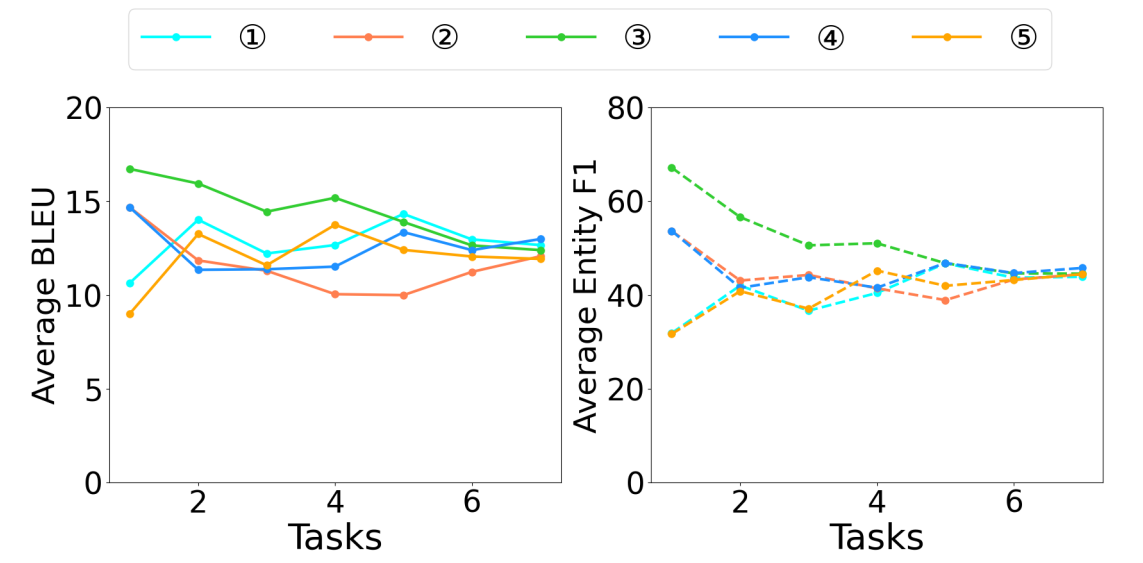}
   \caption{The average results  of TPEM over 7 domains with 5 different orderings randomly sampled from the 7 domains.}
   \label{fig:shuffle}
\end{figure}

\section{Experimental Results}
\paragraph{Main Results}
We evaluate TPEM and baselines with  BLEU~\cite{papineni2002bleu} and entity F1~\cite{madotto2018mem2seq}. 
We conduct experiments by following the common continual learning setting, where experimental data from 7 domains arrives sequentially.
The results of each task are reported after all 7 tasks have been learned. That is, each model keeps learning a new task by using the weights learned from past tasks as initialization. The evaluation results are reported in Table  \ref{table2}. The typical TDSs (i.e., Ptr-Unk, Mem2Seq, GLMP) perform much worse than the continual learning methods (UCL and TPEM). This is consistent with our claim that conventional TDSs suffer from catastrophic forgetting. 
TPEM achieves significantly better results than baseline methods (including Re-init and Re-init-expand) on both new and old tasks. The improvement mainly comes from the iterative network pruning, expanding and masking.

\paragraph{Ablation Study}
To investigate the effectiveness of each component in TPEM, we conduct ablation test in terms of removing  network pruning (w/o Pruning),  network expansion (w/o Expansion), and network masking (w/o Masking). The experimental results are reported in Table \ref{table2}. The performance of TPEM drops more sharply when discarding network pruning than discarding the other two components. This is within our expectation since the expansion and masking strategies rely on network pruning, to some extent. Not surprisingly, combining all the components achieves the best results. Furthermore, by comparing the results of Re-init and Re-init-expand, we can observe that only using network expanding cannot improve the performance of Re-init.

\paragraph{Case Study}
We provide visible analysis on the middle states of all the models. Figure~\ref{fig:task1_7} shows how the results of each task change as new tasks are being learned subsequently. Taking the third task as an example, we observe that the performance of conventional TDSs and UCL starts to decay sharply after learning new tasks, probably because the knowledge learned from these new tasks interferes with what was learned previously. However, TPEM achieves stable results over the whole learning process, without suffering from knowledge forgetting. 

\paragraph{Effect of Task Ordering}
To explore the effect of task ordering for our TPEM model, we randomly sample 5 different task orderings in this experiment. The average results of TPEM over 7 domains with 5 different orderings 
are shown in Figure \ref{fig:shuffle}.  We can observe that although our method has various behaviors with different task orderings, TPEM is in general insensitive to orders because the results show  similar trends, especially for the last 2 tasks.


\section{Conclusion}
In this paper, we propose a continual learning method for task-oriented dialogue systems with iterative network pruning, expanding and masking. Our dialogue system preserves performance on previously encountered tasks while accelerating learning progress on subsequent tasks. Extensive experiments on 7 different tasks show that our TPEM method performs significantly better than compared methods. In the future, we plan to automatically choose the pruning ratio and the number of re-training epochs in the network pruning process for each task adaptively. 

\section*{Acknowledgments}
This work was partially supported by National Natural Science Foundation of China (No. 61906185), Natural Science Foundation of Guangdong Province of China (No. 2019A1515011705), Youth Innovation Promotion Association of CAS China (No. 2020357), Shenzhen Science and Technology Innovation Program (Grant No. KQTD20190929172835662), Shenzhen Basic Research Foundation (No. JCYJ20200109113441941).

\bibliographystyle{acl_natbib}
\bibliography{acl2021}

\end{document}